\documentclass{article}

\usepackage{PRIMEarxiv}

\usepackage[utf8]{inputenc} 
\usepackage[T1]{fontenc}    
\usepackage{hyperref}       
\usepackage{url}            
\usepackage{booktabs}       
\usepackage{amsfonts}       
\usepackage{nicefrac}       
\usepackage{microtype}      
\usepackage{lipsum}
\usepackage{fancyhdr}       
\usepackage{graphicx}       
\graphicspath{{media/}}     

\newcommand\blfootnote[1]{%
  \begingroup
  \renewcommand\thefootnote{}\footnote{#1}%
  \addtocounter{footnote}{-1}%
  \endgroup
}

\title{An active learning model to classify animal species in Hong
Kong}

\author{
  \large{Gareth Lamb, Ching Hei Lo, Jin Wu, Calvin K. F. Lee*} \\
  School of Biological Sciences \\
  University of Hong Kong \\
  Hong Kong\\
  \\}

\begin{document}
\maketitle

\begin{abstract}
Camera traps are used by ecologists globally as an efficient and non-invasive method to monitor animals. While it is time-consuming to manually label the collected images, recent advances in deep learning and computer vision has made it possible to automating this process \cite{chen_deep_2014}. A major obstacle to this is the generalisability of these models when applying these images to independently collected data from other parts of the world \cite{beery_efficient_2019}. Here, we use a deep active learning workflow \cite{norouzzadeh_deep_2021}, and train a model that is applicable to camera trap images collected in Hong Kong.
\end{abstract}


\blfootnote{*Corresponding author}

\section{Introduction}

Wildlife management, conservation, and understanding biodiversity requires accurate monitoring of animals through observation. Camera trap surveys are popular tools used for observing animals due to their non-invasive nature, capacity to sample a broad spectrum of species, and continuous operation throughout the day after they are deployed \cite{burton_wildlife_2015, wearn_mammalian_2017}. While this continuous nature of camera traps offers amazing opportunities for researchers, they also generate vast amounts of data that needs to be processed and analysed, requiring huge investments in time and human efforts. 

Advances in deep learning methods have offered promising alternatives to manual processing of camera trap data, massively reducing time required \cite{norman_can_2023, norouzzadeh_automatically_2018, norouzzadeh_deep_2021, willi_identifying_2019}. Recent advances and developments have led to models that were able to achieve high accuracies when classifying animals to the species level from camera trap images at speeds that are impossible when manually labelling by hand \cite{norman_can_2023, tabak_machine_2019}. 

While the highest accuracies have been achieved when applying these models to data originating from the same region as the training data, accuracies are lower when applied to out-of-sample, or “unseen”, images gathered from other sources \cite{norman_can_2023, tabak_machine_2019}, as these networks do not generalise well. This means that despite the relatively low-cost and effort to set up camera trap projects, these often cannot take advantage of the deep learning models, thus increasing the cost and effort required to fully make use of the collected data \cite{norouzzadeh_deep_2021}. 

One method of overcoming having too few labelled images to train a deep learning model is through via using transfer learning \cite{norouzzadeh_deep_2021}. Transfer learning involves using information from another similar but different task to reduce training costs \cite{iman_review_2023}, improving model accuracies while reducing training time and the amount of labelled data required \cite{hussain_2019, willi_identifying_2019, yosinski_how_2014}. Transfer learning is particularly effective for species classification from camera trap images as existing, openly-available models can provide the information required for general-purpose image classification tasks such as edge detection, while the addition of a smaller, specific target dataset can be used to fine-tune the model 
\cite{hussain_2019, norouzzadeh_deep_2021}. 

Here, we present a model that was trained through transfer learning specifically for camera trap photos collected from Hong Kong. In addition to the model and associated accuracy assessments, we present a pipeline that we aim to provide an easier method for ecologists to adapt to their own study sites. Our pipeline used MegaDetector \cite{beery_efficient_2019, microsoft_megadetector_2023} to first generate cropped images of any animals detected. We then created a model via transfer learning, using a ResNet50 pre-trained from classifying images from the ImageNet database \cite{krizhevsky_imagenet_2012}.

\section{Animal classification workflow}
\label{sec:headings}

Prior to classifying the species of detected animals from camera traps, the images needed to first be sorted, removing any empty images which did not have any animals present within. To achieve this, we used MegaDetector v5a \cite{microsoft_megadetector_2023}, which sorted through all the available images and created bounding boxes for any animals that are present. 

After running all the images through MegaDetector, image crops containing only the animals based on MegaDetector’s outputs are used as input into a species classification model that was trained specifically using training data from Hong Kong.

The data used for this study was gathered by different organisations and groups across Hong Kong. As each group collecting this data had varying objectives, and conducted their data collection independently, the dataset had varying camera placement strategies, camera settings, and camera models. To train the model, we used an initial dataset that included fifteen groupings of animals/species (Table \ref{tab:tab1}).

\begin{table}[h]
    \centering
    \caption{Number of images within each grouping used to classify animals within the dataset.}
    \begin{tabular}{|cc|}
    \hline
        \textbf{Grouping} & \textbf{Count}\\
        \hline
        Birds & 185\\
        \textit{Canis Lupis familaris} & 396\\
        \textit{Lutra lutra} & 1445\\
        \textit{Felis catus} & 80\\
        \textit{Herpestes javanicus} & 71\\
        \textit{Hystrix brachyura} & 3911\\
        \textit{Macaca mulatta} & 1274\\
        \textit{Melogale spp.} & 74\\
        \textit{Muntiacus spp.} & 2733\\
        Other animal & 9\\
        \textit{Paguma larvata} & 165\\
        \textit{Prionailurus bengaliensis} & 1614\\
        Rodent & 185\\
        \textit{Sus scrofa} & 2192\\
        \textit{Viverricula indica} & 2084\\
        \hline
    \end{tabular}
    \label{tab:tab1}
\end{table}

\section{Model development and results}

To create a model that is trained specifically to identify animals in Hong Kong, we first used a transfer learning approach \cite{taylor_transfer_2009}. Taking a pre-trained ResNet50 model that has learned to classify images from the ImageNet database \cite{he_deep_2015}, we finetuned the model using the camera trap images collected in Hong Kong. For this initial model, we were able to achieve an overall accuracy of 89\% using the test set. However, the generalisability of the model is limited, this accuracy drops to 50\% when the model is applied to an independent dataset collected in Hong Kong. 

\begin{figure}
    \centering
    \includegraphics[width=0.75\linewidth]{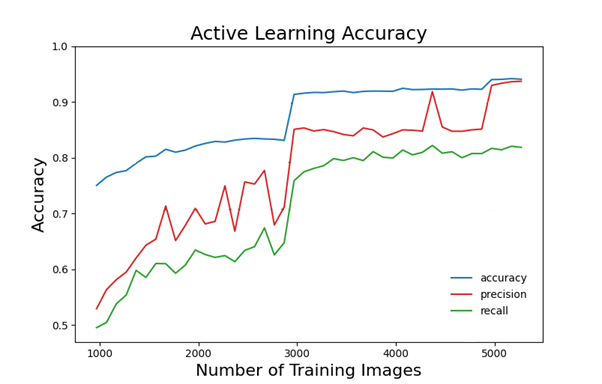}
    \caption{Accuracy, precision and recall of the model as labels are added to training images. }
    \label{fig:fig1}
\end{figure}

A subsequent model was built using an active learning method, following Norouzzadeh et al., \cite{norouzzadeh_deep_2021}. The model achieved incremental improvements to accuracy as training images were added (Fig. \ref{fig:fig1}). Using active learning minimises the time needed to improve the model, as the algorithm is able to specify which images needs labels as the model is being trained. Our pipeline makes use of Timelapse \cite{greenberg_timelapse_2023} to streamline the labelling workflow.

\begin{figure}
    \centering
    \includegraphics[width=0.75\linewidth]{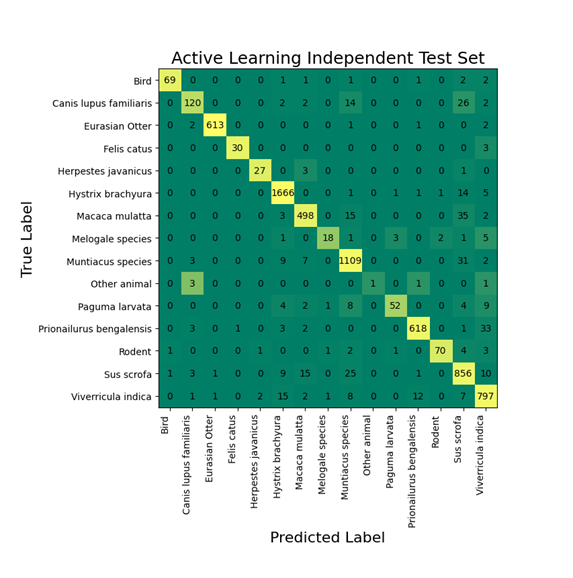}
    \caption{Confusion matrix generated by applying the active learning model onto an independent dataset using images collected from other sites in Hong Kong.}
    \label{fig:fig2}
\end{figure}

To test the generalisability of this active learning model, we subsequently tested the model using an independent dataset using images collected from other sites in Hong Kong, and show the results using a confusion matrix (Fig. \ref{fig:fig2}). 

Based on the independent dataset, the model achieved an overall accuracy of 94.1\%, along with precision, recall, and F1-score of 93.7\%, 81.9\%, and 87.4\%, respectively. Specific classes, such as “other animal”, where the total number of samples available, had much lower accuracies, while more common species achieved F1-scores higher than 90\%. Specifically, “\textit{Melogale spp.}”, “other animals”, and “\textit{Paguma larvata}” had F1-scores lower than 80\%. The model was unable to accurately classify animals into the “other animal” group, as there was only a total of 18 samples within this class. This class was created as a group that included animals there were different from each other (e.g. snakes, tortoises), so no real patterns was detectable within the group. 

\section{Conclusion}

We applied a workflow using MegaDetector and an active learning algorithm to train a deep learning classifier for animals in Hong Kong efficiently and requiring a relatively small amount of labelled images. 

\section{Acknowledgements}

The authors acknowledge Green Power, Kadoorie Farm and Botanical Garden, World Wide Fund Hong Kong, Ecosystems Limited, Aurecon, and Anthony Yim for contributing their camera trap data to the project.

\clearpage

\bibliographystyle{unsrt}  
\bibliography{references}

\end{document}